\documentclass{article}
\usepackage{graphicx} 

\date{October 2024}

\usepackage{amsmath}
\usepackage{amsfonts}
\usepackage{graphicx}
\usepackage{hyperref}

\title{You are out of context!}
\author{Giancarlo Cobino, Simone Farci}

\begin{document}

\maketitle

\begin{abstract}
This research proposes a novel drift detection methodology for machine learning (ML) models based on the concept of "deformation" in the vector space representation of data. Recognizing that new data can act as forces stretching, compressing, or twisting the geometric relationships learned by a model, we explore various mathematical frameworks to quantify this deformation. We investigate measures such as eigenvalue analysis of covariance matrices to capture global shape changes, local density estimation using kernel density estimation (KDE), and Kullback-Leibler divergence to identify subtle shifts in data concentration. Additionally, we draw inspiration from continuum mechanics by proposing a "strain tensor" analogy to capture multi-faceted deformations across different data types. This requires careful estimation of the displacement field, and we delve into strategies ranging from density-based approaches to manifold learning and neural network methods. By continuously monitoring these deformation metrics and correlating them with model performance, we aim to provide a sensitive, interpretable, and adaptable drift detection system capable of distinguishing benign data evolution from true drift, enabling timely interventions and ensuring the reliability of machine learning systems in dynamic environments. Addressing the computational challenges of this methodology, we discuss mitigation strategies like dimensionality reduction, approximate algorithms, and parallelization for real-time and large-scale applications. The method's effectiveness is demonstrated through experiments on real-world text data, focusing on detecting context shifts in Generative AI. Our results, supported by publicly available code, highlight the benefits of this deformation-based approach in capturing subtle drifts that traditional statistical methods often miss. Furthermore, we present a detailed application example within the healthcare domain, showcasing the methodology's potential in diverse fields. Future work will focus on further improving computational efficiency and exploring additional applications across different ML domains.
\end{abstract}

\section{Introduction}

Machine learning (ML) and artificial intelligence (AI) are transforming our world, but their reliance on data creates vulnerabilities. Real-world data is rarely static; it constantly evolves due to shifting contexts, populations, trends, and behaviors. This phenomenon, known as "drift," poses a significant threat to the dependability and accuracy of ML models. Drift takes various forms. \textbf{Concept drift} alters the relationship between input features and target variables, while \textbf{data drift} changes the distribution of input features themselves. \textbf{Context drift}, particularly relevant in conversational AI, occurs when the underlying topic or context of an interaction shifts, potentially leading to incoherent responses and a decline in model performance. 

This research introduces a novel approach, viewing data as points in a high-dimensional vector space and interpreting new data as forces that deform this space. We aim to develop a sensitive and interpretable methodology that captures these deformations, enabling early warnings of performance degradation.

\section{Problem Statement}

Drift can significantly impact the performance and 
reliability of machine learning (ML) models. It leads to a mismatch between the model's original assumptions and the changing realities of the data, resulting in decreased accuracy and potentially erroneous decision-making, especially in critical domains such as healthcare, finance, and autonomous systems. Models that are exposed to drift may produce unreliable outputs, causing users to lose trust. This erosion of trust hinders the adoption of ML systems and necessitates frequent retraining or recalibration, increasing operational costs.

Furthermore, the advancement of artificial intelligence (AI) is closely tied to the models' ability to detect context, seamlessly shift between different contexts, and provide accurate responses accordingly. Without this capability, AI systems may struggle to maintain coherent, contextually relevant conversations or decisions, especially in dynamic environments such as conversational AI or autonomous decision-making systems.

Given these challenges, the need for effective drift detection methodologies becomes paramount. These methodologies provide early warnings, enabling corrective actions before significant model degradation occurs. By detecting the specific type of drift, it is possible to focus retraining efforts, ensuring that updates are both efficient and targeted. Additionally, continuous drift monitoring helps maintain user confidence and ensures that ML models remain adaptable in dynamic environments.

\section{Research Motivation}

Drift detection continues to face challenges despite significant research efforts. Traditional methods effectively detect large distributional shifts but often miss subtle drifts, which can degrade model performance over time. These gradual changes accumulate and damage model accuracy before they become apparent through standard metrics, highlighting the need for more sensitive detection methods capable of signaling drift earlier for timely intervention.

A major limitation of current techniques is their inability to distinguish between types of drift, such as concept drift (changes in the relationship between input and target) and data drift (shifts in input distribution while target relationships remain constant). Without identifying the specific drift type, corrective actions can be inefficient, reducing their effectiveness.

Additionally, many drift detection methods are computationally intensive, limiting their use in real-time applications or resource-constrained environments. This restricts their practicality in systems that require continuous operation.

The vector space deformation approach proposed in this research offers a solution to these limitations. By analyzing data as points in a high-dimensional space and examining how new data deforms this space, subtle drifts that may otherwise go unnoticed can be detected. This method enhances sensitivity, enabling earlier detection of gradual changes that may not affect overall statistical distributions but still impact model performance.

Moreover, this approach is adaptable, as different types of drift manifest as distinct deformations in the vector space, aiding in the identification of specific drift types and supporting more targeted corrective actions. The method’s strong theoretical foundation also paves the way for more robust drift detection techniques beyond traditional statistical methods.

The vector space deformation approach enhances sensitivity by detecting small, gradual deformations in the data's geometric structure, even when overall statistical distributions remain relatively unchanged. By analyzing distinct deformation patterns, our approach helps differentiate between various drift types, such as concept drift or data drift, allowing for tailored corrective actions.

Generative AI models, particularly those used in conversational AI, are highly susceptible to context drift, as subtle shifts in user language or topic can lead to incoherent or irrelevant outputs.

In summary, the vector space deformation approach addresses the key challenges of traditional methods, offering improved sensitivity, adaptability, and computational efficiency. It promises to enhance the reliability and adaptability of machine learning models in dynamic environments, ensuring long-term effectiveness despite changing data.

\section{Literature Review}

\subsection{Statistical Drift Detection}
Statistical methods form a cornerstone of drift detection, often comparing data distributions from different periods or between a baseline dataset and new data. Key techniques include:

\begin{itemize}
    \item \textbf{Population Stability Index (PSI)}: PSI quantifies the change in the distribution of a feature between two samples. It is simple to understand and interpret.
    
    \item \textbf{Kullback-Leibler (KL) Divergence}: Measures the difference between two probability distributions. It is a non-symmetric divergence, meaning that the order in which distributions are compared matters.
    
    \item \textbf{Jensen-Shannon (JS) Divergence}: A symmetric variant of KL divergence, addressing some of its limitations. JS divergence is often more stable when dealing with sparse distributions.
    
    \item \textbf{Other Statistical Distances}: Methods such as the Kolmogorov-Smirnov (KS) test, Wasserstein distance (Earth Mover's Distance), and others provide alternative ways to quantify distributional differences.
\end{itemize}

\subsection{Advantages}
\begin{itemize}
    \item \textbf{Well-Established Foundation}: Statistical methods have a strong theoretical basis and are widely used in data analysis.
    
    \item \textbf{Interpretability}: Measures like PSI and KL divergence provide quantifiable metrics that can be monitored for significant changes.
\end{itemize}

\subsection{Context Drift Detection}

\subsection{Limitations}
Traditional statistical tests often focus on individual features, potentially overlooking complex drifts stemming from interactions between multiple features. Determining appropriate thresholds to distinguish natural data variation from significant drift presents a considerable challenge. Furthermore, many methods rely on assumptions about the underlying data distributions, assumptions that may not always reflect the complexities of real-world data. These limitations can lead to missed detections of subtle yet impactful drifts.

\subsection{Model-Based Approaches}
Model-based drift detection utilizes the outputs or behavior of machine learning models to identify potential drift. Key strategies include tracking how the distribution of model predictions shifts over time, which can signal drift; for example, a sudden increase in uncertain classifications or a change in the proportion of positive versus negative predictions might indicate that the model is encountering data outside its area of expertise. Analyzing changes in the confidence scores or prediction probabilities of a model can also provide valuable clues about drift. A decline in overall confidence, even if predictions remain technically correct, might suggest the model is operating on less familiar data. Adversarial techniques employ 'drift generators' to deliberately create perturbed data samples simulating various drift scenarios. By observing how a model's performance degrades on these 'drifted' inputs, insights into vulnerabilities and potential drift detection thresholds can be gained. Another approach involves training a separate model to distinguish between the original data used to train the primary model and new incoming data. If the discriminator successfully separates the datasets, it strongly suggests that drift has occurred.

\subsection{Advantages}
\begin{itemize}
    \item \textbf{Leverages Existing Models}: Model-based approaches directly utilize the trained machine learning models, potentially reducing the need for additional data collection and analysis.
    
    \item \textbf{Sensitivity to Complex Drift}: Changes in model behavior can sometimes capture subtler drifts that may not be immediately obvious in statistical distributions.
\end{itemize}

\subsection{Limitations}
\begin{itemize}
    \item \textbf{Model Dependence}: The effectiveness of model-based approaches is tied to the quality and representativeness of the original model. A poorly trained model will not offer reliable drift detection signals.
    
    \item \textbf{Interpretability}: Analyzing changes in model outputs may not always directly pinpoint the specific nature of the drift or the features involved.
\end{itemize}

\section{Approach}
\subsection{Data Representation in High-Dimensional Space}
The foundation of our proposed drift detection approach rests on the concept of embedding data points as vectors within a high-dimensional space. To transform raw data into these representations, we carefully consider feature selection and engineering processes. Suitable features might include numerical values (such as age or income), categorical features converted into numerical representations, or features extracted using techniques like image processing or natural language processing (NLP). Each selected feature corresponds to a dimension in the vector space. While simpler datasets may work with only a few dimensions, complex data often requires a very high-dimensional representation to fully capture underlying patterns and relationships.

\subsection{Learning Decision Boundaries and Manifolds}
Within this vector space, machine learning models aim to establish decision boundaries or discover hidden structures. In supervised learning, algorithms search for optimal boundaries (hyperplanes in linear models or more complex shapes in non-linear models) that separate the space into regions corresponding to different classes or outcomes. For example, a support vector machine (SVM) may create a hyperplane to divide data points into two categories. Conversely, unsupervised learning techniques, such as clustering algorithms, work to identify manifolds, which are clusters of data points that exhibit similarities and reside close together in this space.

\subsection{Forces Induced by New Data Points}
We hypothesize that new data points, especially those that differ significantly from the training data, can behave like "forces" acting on the existing vector space representation. To illustrate this, imagine a flexible sheet with data points plotted on it. As new, outlying data points are introduced, they may stretch, compress, or twist the sheet, representing deformations in the vector space. These deformations could manifest in several forms:

\begin{itemize}
    \item \textbf{Shifting}: If the new data exhibits data drift, the overall distribution of points in the space might shift. This can change the distances between clusters or affect the position of decision boundaries learned by the model.
    
    \item \textbf{Stretching or Compressing}: Certain areas of the space may stretch or compress due to changes in the variance or correlations of features in the new data.
    
    \item \textbf{Twisting or Warping}: More complex, non-linear drifts might manifest as twisting or warping of geometric relationships within the vector space, altering the fundamental topology of the data's representation.
\end{itemize}

\section{Mathematical Formulation}
\subsection{Data as Constellation (or Cloud)}
Data should not only be considered as rows and columns, but as points scattered in a vast, multidimensional space. Each dimension in this space represents a feature of the data (e.g., age, income, purchase history, large text, or a generated image). The initial dataset used to train a model forms a "constellation" or cloud in this space, depending on the type of data.

When new data arrives, each new data point acts like a small force. Points that are significantly different from the original data (outliers) exert a stronger pull than those that blend in with the rest. The goal is to detect not only if the center of the data constellation is moving but also how the overall shape is deforming—whether stretching, compressing, or twisting in response to the new data forces.

\section{Steps and Mathematical Formulation}

\subsection{Step 1: Representing Data in Space}
\begin{itemize}
    \item \textbf{Vector Space}: We represent the data points in a vector space, denoted as \( \mathbb{R}^n \), where each data point becomes a vector (an arrow) in this space.
    
    \item \textbf{Embedding}: The process of converting raw data into vectors is crucial. For numerical data, we may use the values directly. For textual data, techniques such as word embeddings (e.g., Word2Vec) can be used to represent words or documents as vectors.
\end{itemize}

\subsection{Step 2: Quantifying the Force of New Data}
For each new data point \( x_i \):
\begin{itemize}
    \item \textbf{Deviation Vector (d)}: We draw an arrow from the center of the original data (often the average, \( \mu \)) to the new point:
    \[
    d = x_i - \mu
    \]
    This gives the direction and magnitude of the "pull" exerted by this new point.
    
    \item \textbf{Force Magnitude}: We can scale the length of the deviation vector \( \|d\| \) to reflect the influence of the new point on the overall shape:
    \begin{itemize}
        \item \textbf{Simple Distance}: Use the length \( \|d\| \) directly.
        
        \item \textbf{Fading Influence}: Use a function that decreases the force as the distance from the center increases, such as:
        \[
        \|d\| \cdot e^{-k\|d\|}
        \]
        where \( k \) controls how quickly the force fades.
        
        \item \textbf{Relative to Spread}: Divide the length \( \|d\| \) by a measure of the spread of the original data (such as its standard deviation), making the force proportional to how "unusual" the new point is.
    \end{itemize}
\end{itemize}

\subsection{Step 3: Capturing the Deformation}
To measure how the entire data space is deforming, we employ the following techniques:

\begin{itemize}
    \item \textbf{Eigenvalues and Eigenvectors (Global Shape)}: Imagine drawing the largest possible ellipse (or ellipsoid in higher dimensions) that encompasses the data points.
    \begin{itemize}
        \item \textbf{Eigenvectors} represent the directions in which the ellipsoid is stretched the most.
        
        \item \textbf{Eigenvalues} provide the magnitude of the stretch in each of those directions.
    \end{itemize}
    By comparing the eigenvectors and eigenvalues of the original data to those of the new data (generated from devices, phones, the internet, etc.), we can calculate whether the data is stretching, compressing, or rotating in significant ways.
    
    \item \textbf{Local Density Estimation (Subtle Shifts)}: Imagine some areas of the data space becoming more crowded with points while other areas thin out.
    \begin{itemize}
        \item We can estimate the density of points in different regions using techniques such as Kernel Density Estimation (KDE).
        
        \item By comparing the density of the new data to the expected density based on the original data, we can identify areas where the data distribution is changing more subtly.
    \end{itemize}
\end{itemize}

\begin{figure}
    \centering
    \includegraphics[width=1\linewidth]{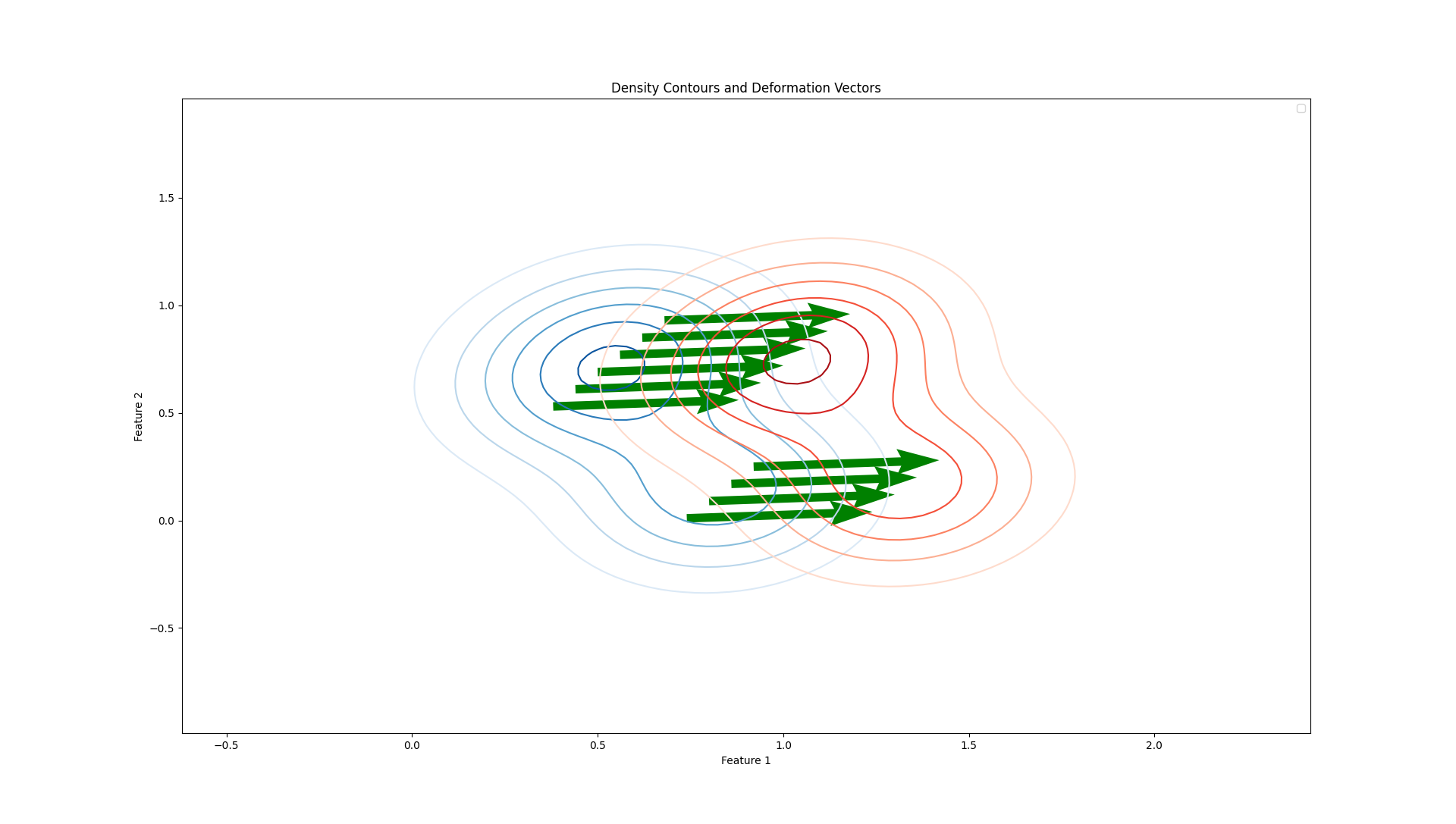}
    \caption{Density contour and transformation (move) vectors}
    \label{fig:enter-label}
\end{figure}
We treat data as vectors in a high-dimensional space, where each feature corresponds to a specific dimension.

\subsection{Deformation as Average Displacement}
A straightforward way to quantify the impact of new data on an existing distribution is by measuring how much, on average, the new points pull the center of the original data. This approach, which we term \textit{average displacement}, relies on the intuitive notion that as new data points arrive, they exert a kind of gravitational force on the overall data cloud, potentially shifting its center of mass.

We can conceptualize this by imagining the initial dataset as a cluster of points held together by elastic bands. Each new data point acts like a tiny weight added to this system, pulling on the bands and potentially shifting the cluster's equilibrium. The average displacement measures the average magnitude of these pulls, calculated as the distance between each new point and the center of the original data (often represented by the mean). A larger average displacement suggests a stronger overall pull away from where the model was initially trained.

While average displacement offers a readily interpretable measure of global shift, it suffers from significant limitations. This approach is akin to tracking only the movement of a ship's anchor. While the anchor's position provides some information about the ship's location, it reveals nothing about the ship's orientation, its rocking due to waves, or whether it is taking on water.

Similarly, average displacement is blind to changes in the shape of the data distribution. A cluster of data points might be stretching, rotating, or becoming more dispersed, and yet the average displacement could remain relatively unchanged. For instance, imagine a scenario where a few outliers in the new data move significantly farther from the center. The average displacement might not change drastically, especially if the dataset is large, because the value is averaged across all data points.

Furthermore, average displacement ignores the direction of these pulls. Two new data points could exert equal but opposite forces on the center of mass, effectively canceling each other out in the average displacement calculation. However, these opposing forces, rather than indicating stability, might be subtly twisting or warping the shape of the data in ways that degrade model performance.

Another interesting concept related to data shape changes is the \textit{convex hull}. Imagine wrapping the data points with the tightest possible elastic sheet. The convex hull represents the boundary of this sheet, encompassing all the data points. A shift in the convex hull often indicates a change in the extremities of the data distribution, where boundary points are pulled outward. While intuitively appealing, relying solely on the convex hull for drift detection poses practical challenges. Computing the convex hull, especially as the number of dimensions increases, becomes computationally expensive, limiting its usefulness for real-time monitoring. Moreover, the convex hull is highly susceptible to outliers. A single data point moving far from the main cluster can drastically alter the convex hull, even if the majority of the data remains relatively stable.

The limitations of average displacement and the computational challenges of the convex hull underscore the need for more sophisticated approaches. To effectively detect drift, we must move beyond simply measuring the average shift in data points towards techniques that capture changes in the spread, orientation, and overall shape of the data distribution within the high-dimensional vector space.

\section{Mathematical Formulation of Deformation as Average Displacement}
Our approach to drift detection is based on interpreting the arrival of new data as a force that deforms the geometric structure of the data space. To make this concept concrete, we model the data in an \( n \)-dimensional Euclidean space \( \mathbb{R}^n \), where each dimension corresponds to a specific feature of the data after appropriate preprocessing and embedding. Each data point is then represented as a vector \( \mathbf{x} = (x_1, x_2, \dots, x_n) \in \mathbb{R}^n \).

\section{Average Displacement: A Basic Measure of Shift}
A preliminary measure of this deformation is the \textit{average displacement}, which captures the overall tendency of new data points to shift the center of the baseline data distribution. Denote the initial (baseline) dataset as \( \{\mathbf{x}_1, \mathbf{x}_2, \dots, \mathbf{x}_m\} \) and the new data points as \( \{\mathbf{x}_{m+1}, \dots, \mathbf{x}_n\} \).

\subsection{Center of Distribution (\( \mu \))}
We first calculate the representative center of the baseline distribution. A common choice is the mean \( \mu \), although the median or a robust estimate may be preferred for data with outliers or non-Gaussian distributions:
\[
\mu = \frac{1}{m} \sum_{i=1}^{m} \mathbf{x}_i
\]
To quantify how the new data points influence the original distribution, we introduce the concept of a force vector. Each new data point \( \mathbf{x}_i \) exerts a pull on the center of the baseline data, analogous to a gravitational force. This force can be represented as a vector \( \mathbf{F}_i \):

\[
\mathbf{F}_i = \mathbf{x}_i - \mu
\]

where:
\begin{itemize}
    \item \( \mathbf{F}_i \) is the force vector exerted by the new data point \( \mathbf{x}_i \).
    \item \( \mathbf{x}_i \) is the vector representing the new data point.
    \item \( \mu \) is the vector representing the center of the baseline data distribution (calculated as the mean, median, or a robust estimate).
\end{itemize}

This force vector captures both the direction and magnitude of the pull. The magnitude, denoted \( \|\mathbf{F}_i\| \), is simply the Euclidean distance between the new point and the baseline center.

\subsection{Average Displacement Calculation}
To assess the overall influence of the new data, we calculate the \textit{average displacement} \( D \), which averages the magnitudes of these force vectors across all new data points:
\[
D = \frac{1}{n - m} \sum_{i=m+1}^{n} \|\mathbf{F}_i\|
\]

where:
\begin{itemize}
    \item \( D \) is the average displacement.
    \item \( n \) is the total number of data points (baseline + new).
    \item \( m \) is the number of data points in the baseline dataset.
\end{itemize}

A larger average displacement \( D \) signifies a stronger overall pull exerted by the new data, indicating a more substantial shift in the data space.

\section{Convex Hull: Capturing the Outer Boundaries}
Another geometric concept relevant to drift detection is the \textit{convex hull}, which encompasses all data points within the smallest possible convex shape. The convex hull of a set of points \( X \) is defined as the set of all possible combinations of those points, where each combination is a weighted average:

\[
\begin{aligned}
\text{Convex Hull}(X) = \left\{ \text{point} \mid \text{point} = \lambda_1 \mathbf{x}_1 + \lambda_2 \mathbf{x}_2 + \dots + \lambda_k \mathbf{x}_k, \right. \\
\left. \mathbf{x}_1, \mathbf{x}_2, \dots, \mathbf{x}_k \in X, \, \lambda_1, \lambda_2, \dots, \lambda_k \geq 0, \, \sum_{i=1}^{k} \lambda_i = 1 \right\}
\end{aligned}
\]

Where:
\begin{itemize}
    \item \( \lambda_1, \lambda_2, \dots, \lambda_k \): These are weights (non-negative numbers) assigned to each data point \( \mathbf{x}_1, \mathbf{x}_2, \dots, \mathbf{x}_k \) in the dataset \( X \).
    
    \item \( \sum_{i=1}^{k} \lambda_i = 1 \): The weights must always sum to 1, ensuring that we are creating a weighted average.
    
    \item \textit{point} \( = \lambda_1 \mathbf{x}_1 + \lambda_2 \mathbf{x}_2 + \dots + \lambda_k \mathbf{x}_k \): This calculates a new point as a combination of the original data points, where each point's contribution is determined by its weight.
    
    \item \( X \) is the set of data points.
\end{itemize}

While the convex hull provides insights into the changing extent of the data distribution, calculating it in high dimensions becomes computationally expensive, making it less practical for real-time drift monitoring.

The convex hull can provide insights into how the distribution of data is changing, especially in terms of its outer limits. A shift in the convex hull indicates that new data is affecting the boundaries of the dataset, possibly signaling a drift. 

Although useful in detecting shifts, calculating the convex hull in high-dimensional spaces is computationally expensive. This makes it less feasible for real-time drift detection, particularly in complex, high-dimensional datasets.

\section{Quantifying Data Deformation: A Multifaceted Approach to Drift Detection}

The success of our proposed drift detection methodology relies on robustly quantifying how new data deforms the geometric structure of the original data space. We model our data within an \( n \)-dimensional Euclidean space \( \mathbb{R}^n \), where each data point, represented as a vector \( \mathbf{x} = (x_1, x_2, \dots, x_n) \in \mathbb{R}^n \), corresponds to a specific combination of features.

\subsection{Average Displacement: A First Step Toward Quantification}
A simple yet intuitive measure of data deformation is \textit{average displacement}. This approach quantifies the overall tendency of new data points to shift the center of mass of the baseline data distribution. Visualize the baseline data as a cloud of points connected by elastic bands. As new data points arrive, they tug on these bands, pulling the cloud in different directions. Average displacement measures the average strength of these pulls.

Let us break down the calculation:

\begin{itemize}
    \item \textbf{Center of Distribution (\( \mu \))}: First, we identify the center of the baseline data, typically represented by the mean \( \mu \). For datasets prone to outliers or with non-Gaussian distributions, the median or a robust estimate might be more appropriate.
    
    \item \textbf{Force Vector (\( \mathbf{F}_i \))}: For each new data point \( \mathbf{x}_i \), we calculate a force vector \( \mathbf{F}_i \). This vector points from the baseline center \( \mu \) to the new point \( \mathbf{x}_i \), representing both the direction and magnitude of the "pull":
    \[
    \mathbf{F}_i = \mathbf{x}_i - \mu
    \]
    The magnitude of this force \( \|\mathbf{F}_i\| \) is the Euclidean distance between the new point and the baseline center.

    \item \textbf{Average Displacement (\( D \))}: To get an overall measure of how much the new data is shifting the baseline distribution, we calculate the average displacement \( D \):
    \[
    D = \frac{1}{n - m} \sum_{i=m+1}^{n} \|\mathbf{F}_i\|
    \]
    where:
    \begin{itemize}
        \item \( D \) is the average displacement.
        \item \( n \) is the total number of data points (baseline + new).
        \item \( m \) is the number of data points in the baseline dataset.
    \end{itemize}
\end{itemize}

\textbf{Higher \( D \)}: A larger value of \( D \) indicates that the new data is pulling strongly on the original data cloud, potentially signaling a significant shift or drift in the underlying data distribution.

\textbf{Lower \( D \)}: A smaller value of \( D \) suggests that the new data is relatively similar to the baseline data, with only minimal shift.
\vspace{1em}

This measure serves as an initial, intuitive step in the drift detection process, helping to detect if new data points are causing substantial changes in the overall data distribution. However, as mentioned earlier, while it gives a basic indication of drift, average displacement doesn't capture changes in the shape or orientation of the data distribution, which may require more sophisticated techniques.

\subsection{Convex Hull: Outlining the Data's Boundaries}
Convex hulls are another geometric subject related to drift detection. Consider the points in your data as pegs on a board. The shape you would get if you wrapped a rubber band around the outermost points is called the convex hull. Formally, it is the smallest convex set that contains all the data points.

The convex hull provides a clear visualization of the data's extent and how this extent changes as new data arrives. A shift in the convex hull suggests that the boundaries of the data distribution are moving.

However, using the convex hull for drift detection presents challenges, especially in high-dimensional spaces. Computing the convex hull becomes increasingly complex as the number of features grows, making it computationally expensive and less practical for real-time monitoring. Additionally, the convex hull is highly sensitive to outliers. A single outlier moving far from the main cluster can drastically alter the convex hull, even if most of the data remains relatively stable.

\subsection{Eigenvalue Analysis: Capturing Shape and Orientation Shifts}
To capture more nuanced deformations, we leverage eigenvalue analysis. The covariance matrix of a dataset summarizes the spread and orientation of the data along different dimensions. By analyzing how the eigenvalues and eigenvectors of the covariance matrix change between the baseline and new data, we can detect stretching, compression, and rotation in the data space.

\begin{itemize}
    \item \textbf{Eigenvalues}: Represent the magnitude of variance along each corresponding eigenvector. Changes in eigenvalues indicate stretching or shrinking of the data distribution along specific dimensions.
    
    \item \textbf{Eigenvectors}: Represent the directions of greatest variance in the data. Shifts in eigenvectors indicate rotations in the data space, suggesting changes in feature correlations.
\end{itemize}

By comparing the eigenvalues and eigenvectors of the baseline and new data covariance matrices, we can quantify these deformations. Two key measures are defined:

\subsection{Eigenvalue Ratio (\( D_{\text{stretch/compress}, i} \))}
\[
D_{\text{stretch/compress}, i} = \frac{\lambda_{Y_i}}{\lambda_{X_i}}
\]
where:
\begin{itemize}
    \item \( \lambda_{Y_i} \) is the \( i \)-th eigenvalue of the new data covariance matrix.
    \item \( \lambda_{X_i} \) is the \( i \)-th eigenvalue of the baseline data covariance matrix.
    \item \( D_{\text{stretch/compress}, i} > 1 \) indicates stretching along the \( i \)-th principal component.
    \item \( D_{\text{stretch/compress}, i} < 1 \) indicates compression along the \( i \)-th principal component.
\end{itemize}

\subsection{Eigenvector Angle (\( D_{\text{rotation}, i} \))}
\[
D_{\text{rotation}, i} = \cos^{-1}\left( \left| \mathbf{u}_{X_i} \cdot \mathbf{u}_{Y_i} \right| \right)
\]
where:
\begin{itemize}
    \item \( \mathbf{u}_{X_i} \) is the \( i \)-th eigenvector of the baseline data covariance matrix.
    \item \( \mathbf{u}_{Y_i} \) is the \( i \)-th eigenvector of the new data covariance matrix.
\end{itemize}
Larger angles indicate more significant rotation in the data space, reflecting changes in the orientation of the data distribution.

\subsection{Local Density Estimation and Comparison: Detecting Subtle Shifts}
While eigenvalue analysis provides a global view of shape changes, local density estimation and comparison allow us to identify subtle drift patterns that might not manifest as drastic changes in the overall shape. Imagine a city's population density map: drift could be represented by new neighborhoods emerging (increased density) or existing ones becoming deserted (decreased density).

We use \textit{Kernel Density Estimation} (KDE) to estimate the probability density function (PDF) of the data, creating a smooth, continuous representation of the "data terrain." By comparing the KDE-estimated density of new data points to the density expected based on the baseline model, we can pinpoint regions of the space where data concentration is changing.

A common way to quantify these density deviations is using the \textit{Kullback-Leibler (KL) divergence}, which measures the difference between two probability distributions. A higher KL divergence indicates a greater difference in local densities, suggesting a potential drift in that region of the data space.

\section{Drawing Insights from Continuum Mechanics: The "Strain" of Data Drift}
To further enrich our understanding of data deformation, we turn to the field of continuum mechanics, which studies the behavior of materials under stress and strain. Imagine stretching a rubber band. The movement of any point on the band from its original to its stretched position is its displacement. Strain, on the other hand, measures the relative deformation within the material—how much the rubber band has stretched compared to its original length.

The key concept here is the relationship between displacement and strain. Strain isn't just about movement; it's about how that movement changes across the material. A high strain indicates a rapid change in displacement over a small distance, like a sharp bend in our rubber band.

We can apply this analogy to our data space. The arrival of new data, particularly those points significantly different from the baseline, can be seen as a force acting on this space, causing "stretching" or "compression" along different dimensions.

\subsection{Strain as a Lens for Drift}
Let's unpack how different types of strain can provide insights into drift:

\begin{itemize}
    \item \textbf{Normal Strain}: This measures the elongation or compression along a specific axis. Imagine a key feature in your model, like "customer age." A significant increase in the average age of new customers would correspond to a stretching of the data distribution along the "age" axis. Normal strain would capture this directly, highlighting features undergoing the most pronounced shifts.
    
    \item \textbf{Lateral Strain}: Things get more interesting when we consider how changes in one feature can ripple through the relationships between other features. Imagine that as customer age increases (stretching along the "age" axis), the feature "income" becomes less correlated with customer behavior. This could manifest as a "squeezing" along the "income" dimension. Lateral strain reveals not just what's changing, but also how these changes affect feature relationships, offering clues about underlying shifts in the data-generating process.
    
    \item \textbf{Volumetric Strain}: This measures the overall expansion or contraction of the data space. A large increase in volume could indicate the emergence of entirely new customer profiles, while shrinking volume might signal that your model is becoming more specialized to a narrower segment of data.
\end{itemize}
\subsection{Formally Defining the Strain Tensor}

\paragraph{Data Representation:}
We represent the data in a metric space $(M, d)$, where $M$ is the data space, and $d$ is the distance metric defined on $M$. For numerical data, this space is typically Euclidean, i.e., $M = \mathbb{R}^n$ with the standard Euclidean distance metric. For more complex data types, such as text or images, the metric space may involve more specialized distance metrics, such as cosine similarity for text embeddings or structural similarity (SSIM) for images. The choice of metric depends on the nature of the data and its structure.

\paragraph{Baseline Data Distribution:}
Let $P$ represent the baseline data distribution, which corresponds to the data used to train the original model. This distribution captures the characteristics of the original dataset and defines the baseline for comparison when analyzing drift.

\paragraph{Displacement Field:}
We introduce the concept of a displacement field, denoted as $\mathbf{v}$. The displacement field is a vector field that captures how new data points displace or "move" points from the original data space. More formally, for each point $x \in M$, $\mathbf{v}(x)$ is the vector that describes the direction and magnitude of the displacement caused by the arrival of new data.

\paragraph{Strain Tensor:}
The strain tensor, denoted as $\boldsymbol{\varepsilon}$, is a mathematical object that captures the local deformation caused by the displacement field at each point in the data space. It is formally defined as the gradient (or derivative) of the displacement field:
\[
\boldsymbol{\varepsilon}(x) = \nabla \mathbf{v}(x)
\]
where $\nabla \mathbf{v}(x)$ is the matrix of partial derivatives of the displacement field $\mathbf{v}$ with respect to the coordinates of $x$. The strain tensor $\boldsymbol{\varepsilon}(x)$ provides a local measure of deformation at the point $x$ in the data space, quantifying how the arrival of new data changes the relationships between neighboring points.

\subsection{Interpreting the Strain Tensor with a Text Example}

\paragraph{Diagonal Elements:}
The diagonal elements of the strain tensor $\boldsymbol{\varepsilon}(x)$ represent the amount of stretching or compression along individual dimensions in the text's word embedding space. Each dimension of the embedding space corresponds to a specific feature of a word’s context or semantic meaning. For example, if we consider a word like "economy," a large positive value along the diagonal element $\varepsilon_{ii}$ might indicate that the word "economy" is appearing more frequently or in a more prominent context in the new data compared to the baseline text. This would suggest a significant shift in the usage of this word, perhaps due to changes in the topics discussed in the new text.

\paragraph{Off-Diagonal Elements:}
The off-diagonal elements of the strain tensor capture changes in the relationships between different words or concepts. These elements reflect how the interactions between words have shifted in the new data. For instance, consider the words "economy" and "inflation." If the off-diagonal element $\varepsilon_{ij}$ between these two words shows a significant change, it could indicate that the relationship between these concepts has evolved in the new text. For example, if "inflation" and "economy" are appearing together more frequently in new contexts, the off-diagonal element would highlight this stronger correlation, reflecting the fact that these two concepts are being discussed in closer relation in the new data.

\subsection{Adapting the Framework to Specific Data Types}
The general strain tensor framework can be tailored to different data types by using data-specific methods for representing and calculating the displacement field and strain tensor. Below, we outline how the framework can be adapted for tabular data, image data, and text data, referencing relevant discussions and sources.

\paragraph{Tabular Data:}
For tabular data, the strain tensor can be calculated by comparing differences in means and covariance matrices between the baseline and new data. The displacement field can be estimated as the difference in feature values between the two datasets, and the strain tensor is then derived from the gradients of this displacement field. The eigenvalues and eigenvectors of the covariance matrix provide insights into how the data is stretched (eigenvalues) and rotated (eigenvectors). This approach is particularly useful for identifying feature-wise changes and shifts in the relationships between features. 

\paragraph{Image Data:}
In the case of image data, pre-trained convolutional neural networks (CNNs) or autoencoders can be employed as feature extractors to represent images as vectors in a lower-dimensional latent space. The strain tensor is then calculated within this feature space, where it captures changes in high-level visual features, such as shapes, textures, or patterns, extracted by the CNN or autoencoder. This approach allows us to analyze how the overall "appearance" of images in the dataset changes over time or between different datasets, by detecting stretching, compression, and deformation of features in this reduced space.

\paragraph{Text Data:}
For text data, changes in word embeddings are analyzed using metrics like cosine distance or Kullback-Leibler (KL) divergence to quantify semantic shifts. Word embeddings represent words as vectors in a semantic space, and the strain tensor is calculated based on changes in pairwise distances between these embeddings. This captures how the relationships between words have changed over time or between different datasets. For example, a significant change in the embedding distances between words like "economy" and "inflation" could indicate a semantic drift, reflecting a shift in the context in which these words are used in the new data. Such metrics provide a way to detect contextual and semantic drift in text data by measuring shifts in the underlying semantic relationships.

\subsection{Relating the Strain Tensor to Specific Types of Drift}

The strain tensor provides a powerful framework for detecting different types of drift in data. By analyzing the diagonal and off-diagonal elements of the strain tensor, we can capture various shifts in the data, whether they relate to concept drift, data drift, or context drift. Below, we relate the strain tensor to each of these drift types.

\paragraph{Concept Drift:}
Concept drift refers to changes in the relationship between features and the target variable over time. The off-diagonal elements of the strain tensor, which capture changes in the correlations between different features, are particularly useful in identifying concept drift. A significant change in the off-diagonal elements indicates that the relationship between features has shifted, which could reflect a change in the underlying target concept. For instance, if the relationship between "age" and "income" has changed in a predictive model for loan approval, this could be a sign of concept drift affecting the decision-making process.

\paragraph{Data Drift:}
Data drift is characterized by changes in the distribution of individual features, such as shifts in their mean or variance. These changes are captured by the diagonal elements of the strain tensor. A shift in the mean of a feature would manifest as a displacement in the corresponding diagonal entry, while an increase in the variance of a feature would be represented as positive strain along that feature's dimension. For example, if a feature such as "temperature" in a weather dataset shows an increase in variance, this would be reflected as a larger positive value in the corresponding diagonal element of the strain tensor, indicating that the distribution of the feature has widened.

\paragraph{Context Drift (Text Data):}
In text data, context drift refers to shifts in the meaning or usage of words within a corpus. Changes in the relationships between word embeddings, as captured by the strain tensor, can signal such contextual shifts. The off-diagonal elements in this case represent changes in the relationships between words. For example, a significant change in the embedding relationship between the words "bank" and "river" versus "bank" and "money" could indicate that the context in which "bank" is used has shifted from a financial meaning to a geographical one, reflecting context drift. This type of drift is particularly important in applications like natural language processing, where understanding the evolving meaning of words is critical to maintaining model performance.

\subsection{Adapting Strain to the Data Landscape}
While the direct physical concepts of strain might not perfectly map onto a high-dimensional data space, the analogy provides a powerful framework. We can mathematically represent the "strain" induced by new data using a \textit{strain tensor}, a mathematical object capable of capturing multi-dimensional changes in distances, angles, and volumes within the data space.

Defining this strain tensor requires carefully considering the specific characteristics of the data. For tabular data, analyzing changes in the covariance matrix can provide a basis for calculating strain. For image data, we might need to work in a lower-dimensional latent space learned by an autoencoder or use image embeddings to represent images as feature vectors.

The key is to adapt the concepts of strain from continuum mechanics to provide a more nuanced and interpretable view of how data is deforming, going beyond simple drift detection to gain a deeper understanding of the nature of the changes and their potential impact on model performance.

\textbf{Note}: While drawing inspiration from continuum mechanics, it is crucial to acknowledge that data space, unlike a physical material, is not truly continuous. There might be scenarios where the direct application of strain-displacement relationships is not appropriate, especially when dealing with discrete data or complex dependencies between features. Further research is needed to refine and validate the applicability of strain-based measures in various data contexts.

\subsection{Limitations and Further Directions}
While these techniques offer valuable insights into data deformation, they also have limitations. Eigenvalue analysis assumes linear relationships between features, which might not hold for all datasets. Local density estimation can be sensitive to the choice of the bandwidth parameter. Moreover, interpreting the results of these methods and relating them to actual model performance degradation requires careful consideration of domain knowledge and the specific characteristics of the application.

Our ongoing research explores more sophisticated techniques, such as \textit{persistent homology} from topological data analysis, to capture even more complex deformations in the data space. We are also investigating the integration of these different measures into a comprehensive

\section{Limitations of Basic Measures}
Both average displacement and the convex hull offer some intuition about data shifts, but they have limitations. 

\begin{itemize}
    \item \textbf{Average Displacement}: This measure is insensitive to changes in the shape, spread, or orientation of the data. It focuses on the global shift but overlooks deformations in the geometric structure of the data.
    
    \item \textbf{Convex Hull}: While useful for detecting boundary changes, the convex hull is highly sensitive to outliers. A single outlier can drastically alter the convex hull, even if most of the data remains stable. Additionally, computing the convex hull in high-dimensional spaces is computationally demanding, limiting its applicability for real-time drift monitoring.
\end{itemize}

These limitations highlight the need for more sophisticated measures that capture the multi-faceted nature of data deformation in a computationally efficient manner.

\subsection{Lack of method for calculating displacement field}
The sources introduce the innovative concept of a displacement field \( \mathbf{v}(x) \), yet they fall short of providing a concrete methodology for its calculation. This omission presents a significant gap in the proposed framework, as the displacement field is the foundation for calculating the strain tensor \( \boldsymbol{\varepsilon}(x) \), which is essential for quantifying data deformation and detecting drift.

Estimating the displacement field in a high-dimensional data space is a non-trivial task. As highlighted in the sources and in our discussion, there is no one-size-fits-all solution; the optimal approach will depend on the nature of the data, the type of drift being examined (concept, data, or context drift), and the computational resources available. Below, we explore potential strategies for estimating the displacement field, drawing on the concepts discussed in the paper.

\subsection{Density-Based Displacement Estimation}

The displacement field can be linked to the density of data points in the new data distribution. The underlying intuition is that points located closer to regions of higher density in the new data are likely to experience larger displacements toward those regions. This can be formalized using density estimation techniques.

\paragraph{Kernel Density Estimation (KDE):}
One method is to estimate the probability density functions (PDFs) of both the baseline data distribution \( P \) and the new data distribution. The displacement vector \( \mathbf{v}(x) \) at a point \( x \) could then be estimated based on the difference between these two PDFs. Specifically, \( \mathbf{v}(x) \) could be modeled as proportional to the gradient of the difference between the new data PDF and the baseline data PDF at \( x \), i.e.,

\[
\mathbf{v}(x) \propto \nabla \left( P_{\text{new}}(x) - P_{\text{baseline}}(x) \right)
\]

This approach aligns with the idea of data drift, where shifts in the density of features are captured by the displacement field, which, in turn, is reflected in the strain tensor. This method captures both local and global changes in the data space, allowing for a detailed understanding of how the new data deforms the original distribution.

\paragraph{Challenges:}
While KDE provides a theoretically sound approach to estimating the displacement field, it is not without challenges. The accuracy of the PDF estimates is highly sensitive to the choice of kernel bandwidth, which must be carefully tuned. Additionally, KDE suffers from the "curse of dimensionality" in high-dimensional spaces, leading to sparse data representation and poor density estimation performance. This issue makes it difficult to apply KDE directly in very high-dimensional spaces, such as those often encountered in modern machine learning applications like image or text data.

\subsection{Manifold Learning and Geodesic Distances}

In cases where the data lies on or near a lower-dimensional manifold within a high-dimensional space, manifold learning techniques can be applied to estimate the displacement field. This approach leverages the underlying structure of the data to provide a more accurate representation of the displacement.

\paragraph{Low-Dimensional Representation:}
The first step is to learn a low-dimensional representation of the data using manifold learning methods such as Isomap, Locally Linear Embedding (LLE), or t-Distributed Stochastic Neighbour Embedding (t-SNE). These methods aim to preserve the intrinsic geometry of the data while reducing the dimensionality, making it easier to estimate displacement and detect drift.

\paragraph{Geodesic Distances:}
Once the manifold is learned, one can compute geodesic distances between points on the manifold. Geodesic distances represent the shortest paths along the surface of the manifold, as opposed to Euclidean distances in the original high-dimensional space. These distances provide a more meaningful measure of how points relate to each other on the manifold's structure.

\paragraph{Estimating the Displacement Field:}
The displacement field \( \mathbf{v}(x) \) could then be estimated by comparing the geodesic distances between corresponding points in the baseline data and the new data. Specifically, by analyzing how the geodesic distances shift between datasets, one can detect and quantify the deformation of the manifold caused by the new data.

\paragraph{Challenges:}
While this approach can be powerful, manifold learning algorithms can be sensitive to noise and parameter choices. In addition, if the data does not lie on a well-defined manifold or the manifold is complex, this approach may not be suitable, as the learned representation might fail to capture the true structure of the data.

\subsection{Alternative Methods for Displacement Estimation}

When estimating the displacement field, there are several alternative methods that can be applied depending on the nature of the data. Below are key techniques that could be utilized:

\paragraph{Dimensionality Reduction Techniques:}
Methods like Principal Component Analysis (PCA) or t-Distributed Stochastic Neighbour Embedding (t-SNE) can be applied first to reduce the data to a lower-dimensional space before estimating the displacement field. This approach helps mitigate the challenges posed by high dimensionality, making density estimation and displacement field calculations more feasible.

\paragraph{Neural Networks for Displacement Estimation:}
Neural networks, specifically autoencoders, can be employed to map data points to a latent space where the displacement field can be learned as a regression problem. The encoder-decoder architecture allows for both dimensionality reduction and displacement estimation. By learning this mapping, the network can capture the underlying structure of the data, providing an interpretable representation of drift.

\paragraph{Flow-Based Models:}
Flow-based models, such as normalizing flows, are another alternative. These models are explicitly designed to estimate the mapping between distributions, making them well-suited for drift detection. Flow-based models can be trained to estimate the displacement field between the baseline and new data directly, offering an interpretable representation of how the data evolves over time.

\subsection{Manifolds}

A manifold is a central concept in geometry and topology, representing a generalized idea of a curve or surface extended to higher dimensions. Formally, a manifold is a topological space that locally resembles Euclidean space near each point. This means that for any given point on a manifold, there is a neighborhood around that point which is homeomorphic to an open subset of Euclidean space. Several key concepts related to manifolds are outlined below:

\begin{itemize}
    \item \textbf{Topological Space:} A manifold is a topological space, meaning it has a topological structure that allows for continuous deformation of subspaces, consideration of limit points, and unions of open sets.
    
    \item \textbf{Homeomorphism:} There exists a homeomorphism between an open subset of the manifold and an open subset of Euclidean space. This implies a local similarity between the manifold and Euclidean space, though the global structure may be much more complex.
    
    \item \textbf{Charts:} A chart is a pair consisting of an open subset of the manifold and a homeomorphism to an open subset of Euclidean space. Charts provide a local coordinate system that simplifies analysis and calculation on the manifold.
    
    \item \textbf{Atlas:} An atlas is a collection of charts that covers the entire manifold. It provides a "map" that allows for transitioning between local coordinate systems, helping to piece together the overall structure of the manifold.
    
    \item \textbf{Differentiable Manifold:} A differentiable manifold is a type of manifold where the transition functions between overlapping charts are differentiable. This structure enables the use of calculus on the manifold and makes it suitable for studying smooth shapes and continuous spaces.
    
    \item \textbf{Riemannian Manifolds:} A Riemannian manifold is a differentiable manifold equipped with a Riemannian metric, which allows for the measurement of distances and angles. This structure is crucial for analyzing curvature and other geometric properties, making it essential for studying smooth and continuous deformations of the data space.
\end{itemize}

Manifolds are particularly useful when the data lies on or near a lower-dimensional subspace within a higher-dimensional space. Techniques such as Isomap or t-SNE can be used to learn the underlying manifold structure. Once the manifold is learned, geodesic distances can be calculated to estimate the displacement field between the baseline and new data points.

\subsection{Step 1: Representing Data in Space}
Data points are represented as vectors in a vector space \( \mathbb{R}^n \), where \( n \) is the number of features. A data point \( x_i \in \mathbb{R}^n \) is a vector in this space. For numerical data, the values can be used directly. For text, we use word embeddings such as Word2Vec to convert text into vectors.

\subsection{Step 2: Quantifying the Force of New Data}
For each new data point \( x_i \), the deviation vector \( d \) is calculated as:
\[
d = x_i - \mu
\]
where \( \mu \) is the center (mean) of the original data distribution. The magnitude of the deviation can be scaled to reflect the influence of the new data point. We can scale the length of the deviation vector using a fading influence function:
\[
\text{Force Magnitude} = \| d \| e^{-k \| d \|}
\]
where \( k \) controls how quickly the influence fades.

\subsection{Covariance Matrix and Eigenvalue Analysis}

In physical deformation, strain measures how much a material is stretched or compressed in different directions. In a data space, stretching or compressing can be quantified through changes in the covariance matrix of the dataset, where the covariance matrix captures how features (dimensions) of the data vary with one another.

\textbf{Covariance Matrix:} In a high-dimensional dataset, the covariance matrix $\Sigma$ describes the pairwise covariances between all features:

\[
\Sigma = \frac{1}{N-1} \sum_{i=1}^{N} (x_i - \mu)(x_i - \mu)^T
\]
where $\mu$ is the mean vector of the features, and $x_i$ represents a data point in the feature space.

\textbf{Eigenvalue Analysis:} The eigenvalues of the covariance matrix describe the variance along the principal axes of the data. If a new batch of data causes certain eigenvalues to grow or shrink significantly, this corresponds to a "stretching" or "compressing" of the data along those directions. This provides a precise, quantifiable measure of deformation:

\begin{itemize}
    \item \textbf{Stretching:} An increase in variance (eigenvalue) along certain dimensions.
    \item \textbf{Compressing:} A decrease in variance along certain dimensions.
\end{itemize}

\textbf{Eigenvectors:} The eigenvectors correspond to the directions in the data space where the variance is the greatest. Changes in these directions (e.g., rotation of the eigenvectors) would signify a "twisting" of the data space. This captures non-linear, complex drifts that affect feature relationships.

\textbf{Comparing Deformations:} You can compare the covariance matrix of the new data to the baseline (training) data:

\[
D_{\text{stretch/compress}, i} = \frac{\lambda_{\text{new}, i}}{\lambda_{\text{old}, i}}
\]
where $\lambda_{\text{new}, i}$ and $\lambda_{\text{old}, i}$ are the $i$-th eigenvalues of the new and old data covariance matrices, respectively. A value greater than 1 indicates stretching, while a value less than 1 indicates compression.

\subsection{Local Density Estimation and Deformation}

In cases where drift manifests as subtle shifts in the distribution (i.e., local clusters of data become denser or sparser), traditional methods like covariance matrices might not capture these deformations. Instead, local density estimation can help quantify more nuanced shifts.

\textbf{Kernel Density Estimation (KDE):} KDE is a non-parametric way to estimate the probability density function (PDF) of the data. You can use KDE to detect local shifts in data density, which can correspond to regions of the space where data is "compressing" or "expanding" due to new data.

The local density difference between the old and new data can be expressed as:

\[
\Delta \rho(x) = \hat{\rho}_{\text{new}}(x) - \hat{\rho}_{\text{old}}(x)
\]
where $\hat{\rho}_{\text{new}}(x)$ and $\hat{\rho}_{\text{old}}(x)$ represent the local density estimates at point $x$ for the new and old data, respectively.

\textbf{Kullback-Leibler Divergence:} To quantify how significantly the local densities differ, you can use Kullback-Leibler (KL) divergence:

\[
KL(P \| Q) = \sum P(x) \log \frac{P(x)}{Q(x)}
\]
where $P(x)$ is the PDF of the old data, and $Q(x)$ is the PDF of the new data. Large KL divergence would signal significant local deformations, which can indicate drift that traditional global metrics might miss.

This connects back to the metaphor by identifying regions of the data space where clusters become more tightly packed (compression) or dispersed (expansion), echoing the physical concept of deformation.

\subsection{Addressing Deformation in Complex Data Types (Text and Categorical Data)}

Not all data is numerical or continuous, and here the deformation metaphor can face challenges. Let’s explore how this idea can be applied to non-numerical data types:

\textbf{Textual Data:} In the context of text data, embeddings (e.g., Word2Vec, BERT embeddings) transform words, phrases, or documents into vectors in a high-dimensional space. The "semantic space" of word embeddings allows you to measure distances between words and phrases in a meaningful way.

\begin{itemize}
    \item \textbf{Stretching in Text:} If certain words in a corpus become semantically closer to each other (due to changes in context), this can be seen as a "compression" of the semantic space in those regions.
    \item \textbf{Twisting in Text:} Changes in word relationships (e.g., "bank" shifting its primary context from finance to geography) can be represented as a rotation in the space of word embeddings.
\end{itemize}

\textbf{Categorical Data:} For categorical features, distances between categories may not be straightforward. However, you can represent categories using one-hot encoding or embedding layers (e.g., using embeddings for categorical features as in neural networks). Then, similar techniques like covariance or density estimation can be applied in the transformed vector space.

While the deformation analogy is effective for continuous data representations like numerical or image-based datasets, it may be less intuitive for categorical or highly discrete data types. The continuous structure assumed in deformation metrics does not perfectly map onto non-continuous data, and therefore, interpretations of stretching or compressing may be more abstract in these cases.

In both cases, the underlying idea of deformation remains the same: you're looking for shifts in the relationships between data points that can be quantified geometrically, even though the data itself may be abstract.

\section{Results of Drift Analysis}
The custom drift detection and analysis system, which employs both deformation-based and statistical measures, was tested on real text data. The system was built using the following key methods:

\subsection{Deformation (Cosine Distance)}
This metric measures the semantic change between the original and drifted texts. The cosine distance captures the shift in vector representations of the texts, where a value closer to 0 indicates minimal deformation, and values closer to 1 suggest a significant semantic drift. This value indicates a moderate semantic drift between the original and drifted text. It highlights that the drifted text’s content has shifted contextually or semantically from the original, despite the statistical distribution of words potentially remaining similar.

\subsection{Shape Change (L2 Norm)}
This metric captures the geometric shift in the high-dimensional vector space of the original and drifted texts. The L2 norm measures how the structure and relationships between words have changed. A higher value suggests significant deformation in the usage patterns and associations between words. This indicates a substantial change in how the words and concepts in the drifted text are used compared to the original text. This large value suggests that while the topics might seem familiar, the way they are expressed and structured has notably shifted, supporting the presence of a deeper drift than what basic frequency shifts might capture.

\subsection{Wasserstein Distance}
This is a statistical measure used for comparison. The Wasserstein distance measures the difference between the word distributions of the original and drifted texts. This is a common method for identifying statistical drift by comparing how the probability distribution of words has changed. A relatively low value here indicates that there is little change in the overall distribution of words. However, this suggests that the statistical drift alone would imply minimal changes, which contrasts with the results from the deformation-based analysis.

\section{Experimental Results}

Our drift detection and analysis system was tested on real text data, focusing on contextual and semantic shifts. Using deformation-based metrics (Cosine Distance, L2 Norm) alongside statistical measures (Wasserstein Distance), we were able to detect context drift that was missed by statistical approaches alone. 

The results highlighted the practical benefits of the deformation-based approach, offering a more nuanced understanding of drift in dynamic environments such as conversational AI.

\section{Deformation vs. Statistical Drift}
The Wasserstein distance implies minimal statistical drift, indicating that the distribution of words between the original and drifted text remains largely similar. However, the deformation-based measures (Cosine Distance and L2 Norm) provide a more nuanced picture. The significant Deformation (Cosine Distance) and Shape Change (L2 Norm) reveal that, while the word distributions may not have drastically shifted, the context and relationships between words have changed substantially, indicating contextual or semantic drift that is missed by traditional statistical methods.

\section{Computational Complexity}

The proposed deformation-based drift detection methodology provides enhanced sensitivity and flexibility in identifying subtle shifts in data distributions. However, the computational cost of some of the core operations, particularly in high-dimensional spaces, can become prohibitive, especially for real-time applications or very large datasets.

\subsection{Eigenvalue and Eigenvector Calculations}

\textbf{Challenge}: Eigenvalue and eigenvector analysis is essential for quantifying the shape and orientation of data distributions. However, calculating the eigenvalues and eigenvectors of covariance matrices in high-dimensional spaces can be computationally expensive, particularly as the number of features grows. In large datasets or applications requiring real-time drift detection, this might result in delays or require significant computational resources.

\textbf{Mitigation Strategies}:
\begin{itemize}
    \item \textbf{Dimensionality Reduction}: One way to reduce the complexity is to first apply \textit{Principal Component Analysis (PCA)} or other dimensionality reduction techniques to project the data into a lower-dimensional space before performing eigenvalue analysis. This can preserve much of the structure of the data while significantly reducing the computational burden.
    \item \textbf{Randomized Algorithms}: Another option is to use \textit{randomized algorithms} for eigenvalue decomposition, which can approximate the dominant eigenvalues and eigenvectors with lower computational cost, sacrificing a small degree of precision for a significant speedup.
\end{itemize}

\subsection{Convex Hull Calculation}

\textbf{Challenge}: The convex hull, which provides insights into boundary shifts in the data distribution, becomes increasingly computationally expensive to compute as the number of dimensions increases. In high-dimensional spaces, constructing the convex hull requires evaluating many facets, which scales poorly with data dimensionality.

\textbf{Mitigation Strategies}:
\begin{itemize}
    \item \textbf{Approximation Techniques}: Rather than calculating the exact convex hull, \textit{approximate algorithms} (e.g., Quickhull or iterative pruning) can provide a good-enough estimate of the convex hull in significantly less time. These methods focus on identifying key boundary points rather than constructing the entire hull, thus reducing computational complexity.
    \item \textbf{Sampling}: Subsampling the dataset is another potential strategy. By selecting a representative subset of data points, you can compute an approximate convex hull that reflects the overall shape of the data without requiring full-scale computation. This tradeoff between accuracy and computational load can be useful in real-time settings.
\end{itemize}

\subsection{Kernel Density Estimation (KDE)}

\textbf{Challenge}: KDE is a powerful tool for detecting subtle shifts in local data density, but its performance can degrade in high-dimensional spaces due to the curse of dimensionality. KDE requires careful bandwidth selection, and high-dimensional datasets often lead to sparse data points, making density estimation less accurate and more computationally intensive.

\textbf{Mitigation Strategies}:
\begin{itemize}
    \item \textbf{Dimensionality Reduction}: Similar to eigenvalue analysis, dimensionality reduction techniques such as PCA can be employed before KDE to reduce the number of dimensions, improving both the accuracy and efficiency of the density estimation.
    \item \textbf{Fast Approximation Algorithms}: There are also fast approximation methods for KDE, such as \textit{FFT-based methods} or \textit{tree-based algorithms} (e.g., KD-trees), which can significantly reduce the computational burden by optimizing the kernel bandwidth and data structure.
\end{itemize}

\subsection{Real-Time Feasibility}

\textbf{Challenge}: For systems requiring real-time drift detection (e.g., autonomous systems, financial markets, or conversational AI), the time required for full-scale deformation analysis may be prohibitive. A balance must be struck between sensitivity and computational feasibility.

\textbf{Mitigation Strategies}:
\begin{itemize}
    \item \textbf{Incremental Algorithms}: Instead of recalculating deformation metrics from scratch with each new data batch, \textit{incremental methods} can be used. These methods update eigenvalue decompositions, covariance matrices, and density estimates incrementally as new data arrives, reducing the need for full re-computation.
    \item \textbf{Parallelization and GPU Acceleration}: Many of the computations required for deformation analysis, such as eigenvalue decomposition and convex hull computation, can be parallelized. Leveraging \textit{GPU acceleration} or distributed computing frameworks (e.g., Spark or Dask) can improve performance in large-scale, real-time applications.
\end{itemize}

\subsection{Scalability for Large Datasets}

\textbf{Challenge}: As the size of the dataset grows, the computational load of analyzing deformations can increase significantly, both in terms of memory and processing power. High-dimensional datasets, in particular, can lead to performance bottlenecks.

\textbf{Mitigation Strategies}:
\begin{itemize}
    \item \textbf{Batch Processing}: Instead of processing the entire dataset at once, \textit{batch processing} can be used to handle data in chunks, allowing the system to process data in a more memory-efficient way while still detecting drift in near real-time.
    \item \textbf{Streaming Approaches}: For continuous data, \textit{streaming approaches} allow the system to process data incrementally and detect drift as it occurs, without the need to store and process the entire dataset in memory. This can be particularly useful for applications such as real-time monitoring of data streams.
\end{itemize}

\section{Practical Results}
We have performed several tests, and we invite you to do the same. We used a publicly available text file. Using different language models, we created various models, modifying words and entire sentences to simulate potential drift.

Comparing the original file with one of the synthetic versions, we obtained the following results:

\begin{table}[h!]
\centering
\begin{tabular}{|p{3cm}|p{2cm}|p{5cm}|}
\hline
\textbf{Metric} & \textbf{Value} & \textbf{Description} \\ \hline
Original Text Length & 16,380 characters & Length of the original text \\ \hline
Drifted Text Length & 9,012 characters & Length of the drifted text \\ \hline
Length Change & 45.0\% & Percentage reduction in text length \\ \hline
Deformation (Cosine Distance) & 0.1516 & Measure of overall semantic change (0-1 scale, higher values indicate more change) \\ \hline
Shape Change (L2 Norm) & 0.5507 & Measure of change in specific word usage and frequency \\ \hline
Wasserstein Distance & 0.0047 & Statistical measure of change in word distribution \\ \hline
\end{tabular}
\caption{Drift Analysis Results}
\end{table}

From a visualization perspective, we showed the change in dimensionally reduced space and examined how it evolved. We captured several snapshots of the deformation process. Running the code allows you to see all the deformation steps dynamically.
In the appendix you will find the results of space deformation.

\subsection{Space Deformation}
The results are visible, reduced in dimensionality, as convex hulls for each vector space—original and drifted—with the forces applied to the points, which reshape the original space into the drifted one.

\begin{figure}
    \centering
    \includegraphics[width=1\linewidth]{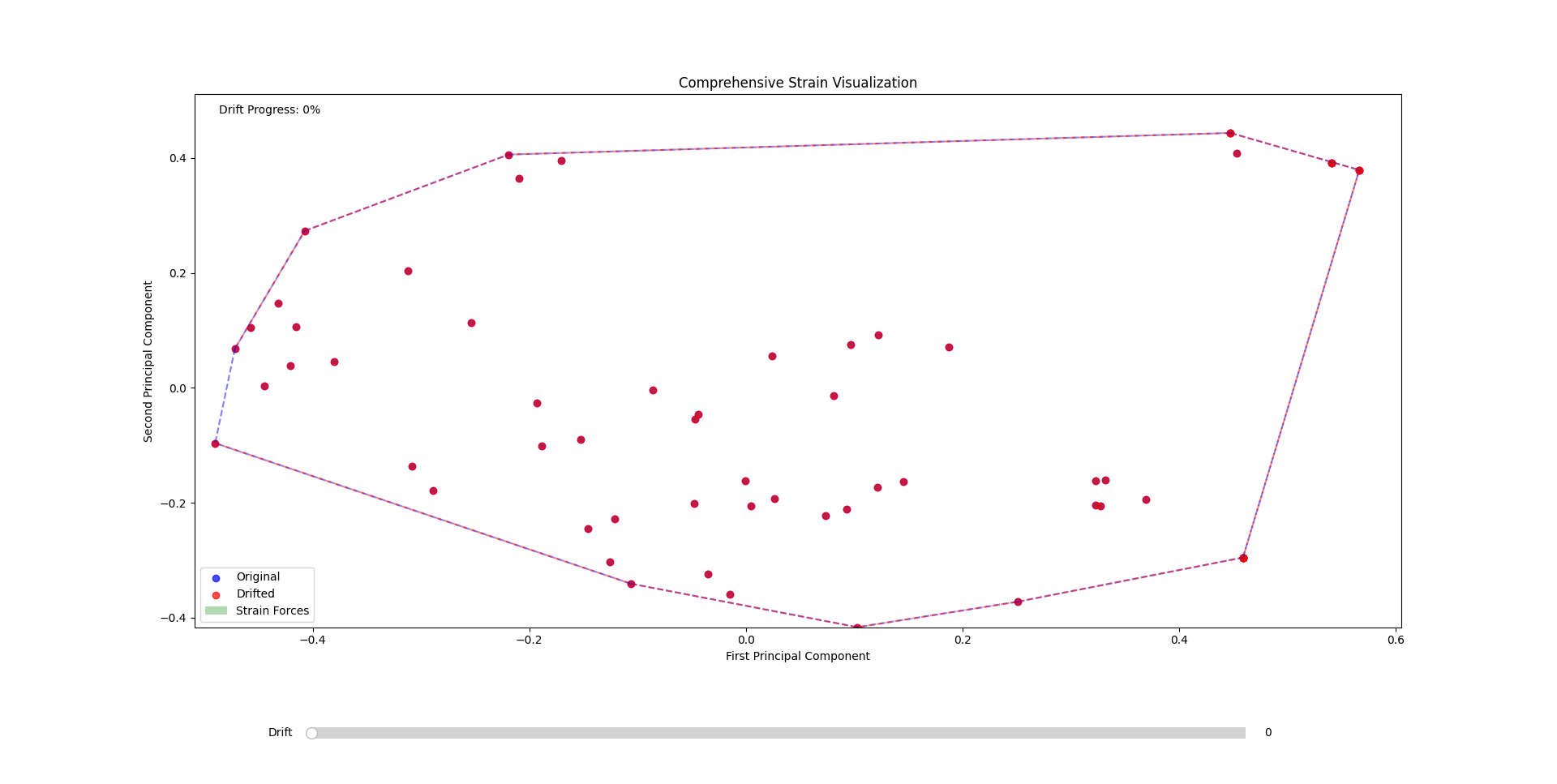}
    \caption{Spaces at time 0 are the same, as new text has not shifted the original}
    \label{fig:vector-space-deformation}
\end{figure}
\begin{figure}
    \centering
    \includegraphics[width=1\linewidth]{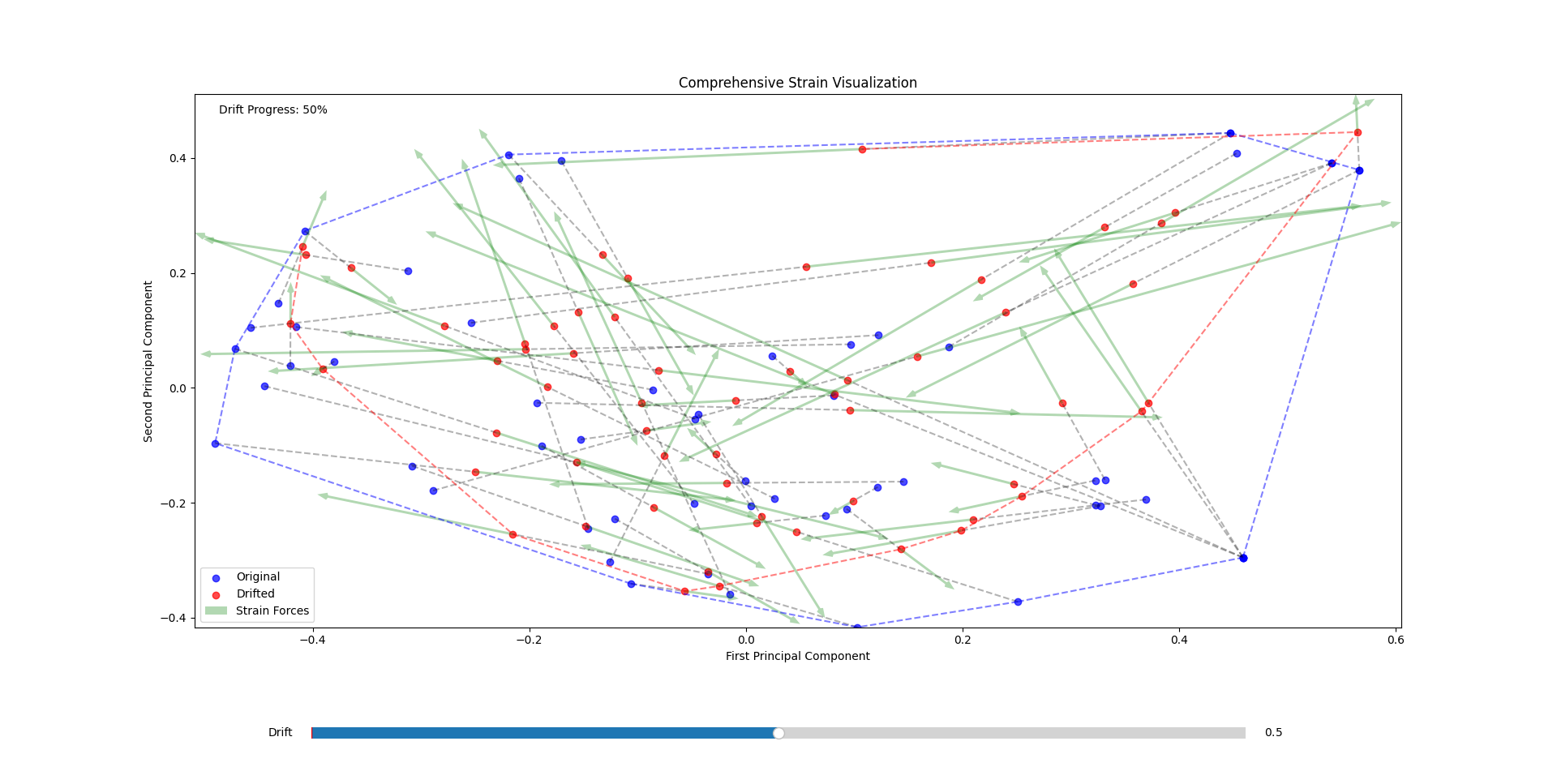}
    \caption{Original space at 50pc is deformed, forces (in green) are strong}
    \label{fig:vector-space-deformation}
\end{figure}
\begin{figure}
    \centering
    \includegraphics[width=1\linewidth]{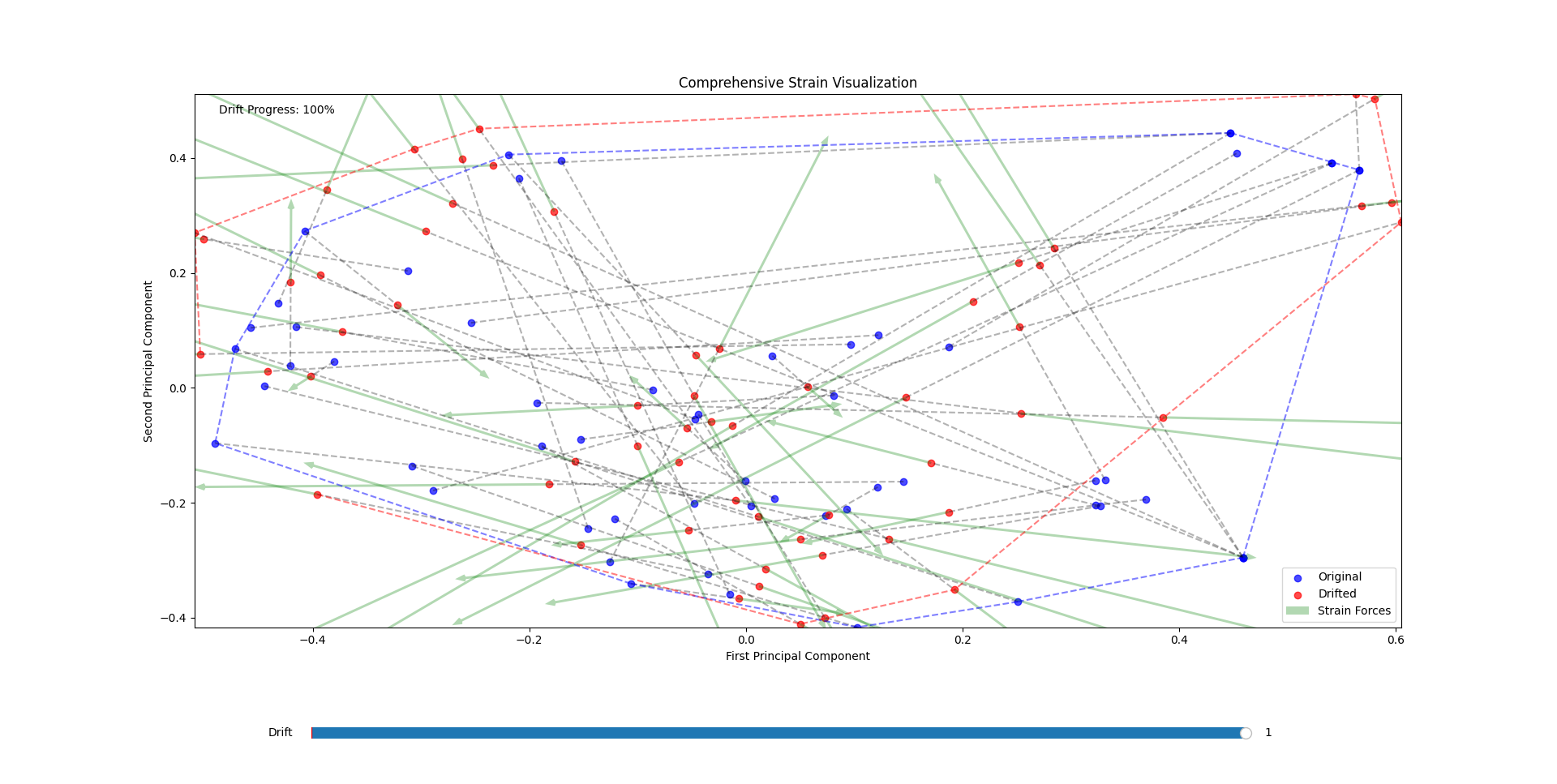}
    \caption{Original space at full force applied has shifted, compress somewhere, expanded in other parts}
    \label{fig:vector-space-deformation}
\end{figure}
\clearpage 

\section{Fields of Application}

The presented model has numerous specific applications in contexts that are already prepared for its use and integration. Similar models have previously benefited from machine learning, demonstrating their potential effectiveness. The idea behind this hypothesis is to demonstrate that the approach is not confined to a specific type of data (such as textual), but can be applied in a wide range of sectors with different dynamics and logics. This would exponentially increase the value and impact of the methodology, showing its versatility and adaptability.

\section{Healthcare}
In medical and healthcare scenarios, we have several successful cases of ML models where a lot of historical data has been used to predict and create dedicated models in a predictive way. But as we know, patient data, symptoms and pathologies can constantly change. As well as the reactions of symptoms between patients that are constantly changing, such as health status, aging and recovery from surgery, rehabilitation and post-operative course.

Changing lifestyles of patients: like time, the behavior of an entire population can change (new treatments, demographic and birth rate changes, new drugs tested that had not been invented before). These changes could make models trained with historical data obsolete. Our “deformation” approach could be used to detect when the data of a patient or group of patients starts to deviate from the reference population on which the model was trained.
The example we tested is the following: Imagine a predictive model used to diagnose a cardiovascular disease. If the patient population starts to age (higher incidence of elderly patients than when the model was trained), there could be a deformation in the vector space that signals a drift. This would allow us to intervene by updating the model before it starts to lose accuracy.

\textbf{Mathematical formulation of drift and deformation}
Data Vector Space Patient data can be represented as vectors in a high-dimensional space. Each vector represents a patient with his or her characteristics:
\[
xi=[xi1 \ xi2 \ xin]^t
\]

Where  xi represents a single patient and each xij is a feature of the patient (e.g. age, cholesterol, blood pressure, etc.). The total data space is then $R^n$, 
where nnn is the number of features

We define a mean $\mu_0$ and a covariance $\Sigma_0$ to describe the distribution of the original model data (patients with a mean age and distribution of other clinical characteristics):

\[
\mu_0 = \frac{1}{N} \sum_{i=1}^{N} x_i, \quad \Sigma_0 = \frac{1}{N-1} \sum_{i=1}^{N} (x_i - \mu_0)(x_i - \mu_0)^T
\]

Where \textit{N} is the number of patients in the original training set.

From a health perspective, over time, new patients with slightly different characteristics are added, both for the same hospital, medical practice and also on a larger sample for example if analyzed at population level (e.g. older population in a given territorial band). We therefore define the mean and covariance of the new data:

\[
\mu_t = \frac{1}{M} \sum_{i=1}^{M} x_{i, \text{new}}, \quad \Sigma_t = \frac{1}{M-1} \sum_{i=1}^{M} (x_{i, \text{new}} - \mu_t)(x_{i, \text{new}} - \mu_t)^T
\]

Where \textit{M} is the number of new patients entering the model.
As already described, "deformation" can be measured and calculated as a difference between the original data distribution and the new data distribution. One way to do this is to analyze the distance between the means and the difference in the covariance.

Distance between means (Global shift): The distance between the original mean and the new data mean can be measured using the scientifically recognized Euclidean norm for similar cases:

\[
D\mu = \|\mu_t - \mu_0\|
\]

A high value of D$\mu$
indicates that the new data is "moving towards" the center of the distribution compared to the original data. This could indicate an increase in the average age of the patients, for example, or a different reaction for the same sample analyzed.

The difference between the covariance could be measured using the Frobenius norm or the comparison of the eigenvalues of the two covariance matrices. Thus, exactly the same properties as vector norms are recognized; this reflects the fact that the matrix space is isomorphic to the vector space for example by the application that sends a matrix into the vector containing its rows one after the other and therefore a matrix norm must have at least the same properties as a vector norm. The Frobenius norm therefore quantifies the global change in the shape of the distribution:
\[
D\Sigma = \|\Sigma_t - \Sigma_0\|_F
\]

Where the Frobenius norm is defined as:
\[
\|\Sigma_t - \Sigma_0\|_F = \sum_{i,j} (\Sigma_t(i,j) - \Sigma_0(i,j))^2
\]

A high value of  D$\sigma$
 would indicate that the shape of the data distribution is changing due to factors that were not initially present or were not considered correctly, suggesting that the new patients being analyzed differ substantially from the original group (for example, a change in the association between cholesterol elevation and the age at which it occurred).

In this analyzed scenario we can define a composite drift index that combines both the shift of the mean of the sample taken as analysis and the change in the covariance to have an overall measure of the drift of all the elements of the function:

\[
D_{\text{totale}} = \alpha D_{\mu} + \beta D_{\Sigma}
\]

Where $\alpha$ and $\beta$ are weights that can be adjusted depending on the relative importance of the mean shift and the shape of the distribution.

From a medical point of view, it is possible to imagine a threshold from a temporal point of view that adapts the drift when it starts to be significant and relevant from the point of view of the tolerance threshold.

The threshold \textit{T} in this case could be set to change when it becomes significant.
If $D_{\text{total}} > T$, it indicates a significant drift.

When the drift exceeds this medical tolerance threshold, the model must be retrained or adapted to account for new data entered on new factors previously not used for the study.

Original population: Mean age  $\mu_0 = 55$, covariance of characteristics (age, cholesterol, blood pressure, etc.).
New population: Mean age $\mu_t = 70$ years. The distance $D_{\mu}$ would be high, indicating that the new patients are significantly older.
Changed covariance: The association between age and cholesterol may be different in the new population, which would result in an increase in $D_{\Sigma}$, indicating that the relationship between the variables is changing.

As you can imagine, this type of approach has several applications, for example, the world of finance could use the model by predicting the risk of default on corporate loans based on macroeconomic variables such as GDP, inflation, unemployment. Before the crisis, the means and covariances between these variables are stable. During a recession, however, unemployment increases and becomes more correlated with the risk of default. To understand the rates of employment and unemployment, for example, to understand the change in covariance, the relationship between the inflation rate and unemployment could change drastically during a possible crisis.
The model in this case can detect when the data begins to "deform", indicating that the model may no longer be valid and needs to be updated.

This type of approach has several applications, for example, the world of finance could use the model by predicting the risk of default on corporate loans based on macroeconomic variables such as GDP, inflation, unemployment. Before the crisis, the means and covariance between these variables are stable. During a recession, however, unemployment increases and becomes more correlated with the risk of default. To understand the rates of employment and unemployment, for example, to understand the change in covariance, the relationship between the inflation rate and unemployment could change drastically during a possible crisis.
The model in this case can detect when the data begins to "deform", indicating that the model may no longer be valid and needs to be updated.

\section{Conclusion}

We introduced a novel approach to drift detection by conceptualizing changes in data as deformations in a high-dimensional vector space. Using mathematical tools from linear algebra, topology, and continuum mechanics, we developed a sensitive and interpretative framework for detecting drift, including the nuances of context drift. Our approach captures both global and local shifts, distinguishing between benign data evolution and performance-degrading drifts. The method has proven effective in dynamic environments, particularly in applications such as conversational AI.

Future work will focus on improving computational efficiency in high-dimensional spaces and expanding the application of this methodology across different ML domains.

\section{Bibliography}

\begin{itemize}
    \item Kullback, S., \& Leibler, R. A. (1951). On information and sufficiency. *The annals of mathematical statistics*, *22*(1), 79-86. (This is the foundational paper on KL Divergence. More recent works applying KL Divergence to drift detection would be needed for a complete reference)
    \item Lin, J. (1991). Divergence measures based on the Shannon entropy. *IEEE transactions on information theory*, *37*(1), 145-151. (This paper discusses JS Divergence and its properties). Again, application to drift detection needs further specification.
    \item Darling, D. A. (1957). The Kolmogorov-Smirnov, Cramér-von Mises tests. *The Annals of Mathematical Statistics*, *28*(4), 823-838. (KS Test)
    \item Villani, C. (2008). *Optimal transport: old and new*. Springer Science \& Business Media. (Wasserstein Distance – A comprehensive text, specific application to drift detection needs additional sources).
    \item Mikolov, T., Sutskever, I., Chen, K., Corrado, G. S., \& Dean, J. (2013). Distributed representations of words and phrases and their compositionality. *Advances in neural information processing systems*, *26*.
    \item Cortes, C., \& Vapnik, V. (1995). Support-vector networks. *Machine learning*, *20*(3), 273-297. (Foundational SVM paper. More recent applications within drift detection would be beneficial).
    \item Tenenbaum, J. B., De Silva, V., \& Langford, J. C. (2000). A global geometric framework for nonlinear dimensionality reduction. *science*, *290*(5500), 2319-2323. (Isomap)
    \item Roweis, S. T., \& Saul, L. K. (2000). Nonlinear dimensionality reduction by locally linear embedding. *Science*, *290*(5500), 2323-2326. (LLE)
    \item Maaten, L. v. d., \& Hinton, G. (2008). Visualizing data using t-SNE. *Journal of machine learning research*, *9*(Nov), 2579-2605. (t-SNE)
    \item Gama, J., Žliobaitė, I., Bifet, A., Holmes, G., \& Žliobaite, I. (2014). A survey on concept drift adaptation. *ACM computing surveys (CSUR)*, *46*(4), 1-37. (A broad survey on concept drift)
 
\end{itemize}

\end{document}